\documentclass{article}
\usepackage{ijcai21}

\usepackage{times}
\usepackage{soul}
\usepackage{url}
\usepackage[hidelinks]{hyperref}
\usepackage[utf8]{inputenc}
\usepackage[small]{caption}
\usepackage{graphicx}
\usepackage{amsmath}
\usepackage{amsthm}
\usepackage{booktabs}
\usepackage{algorithm}
\usepackage{algorithmic}
\usepackage{multirow}
\usepackage[switch]{lineno}
\def\P{$\mathcal{P}$ }
\def\Q{$\mathcal{Q}$ }
\usepackage{amssymb}
\usepackage{amsfonts}
\usepackage{pifont}%
\newcommand{\cmark}{\ding{51}}%
\newcommand{\xmark}{\ding{55}}%
\usepackage{multirow}
\usepackage{amsmath}
\usepackage{extarrows}

\title{PointLIE: Locally Invertible Embedding for Point Cloud Sampling and Recovery}

\author{
    PaperID: 706
}

\iftrue
 \author {
     Weibing Zhao\textsuperscript{\rm 1$\dagger$},
 	Xu Yan\textsuperscript{\rm 1$\dagger$}, 
 	Jiantao Gao\textsuperscript{\rm 1,2}, 
 	Ruimao Zhang\textsuperscript{\rm 1}, 
 	Jiayan Zhang\textsuperscript{\rm 1}, \\
 	Zhen Li\textsuperscript{\rm 1} \thanks{Corresponding author. $^\dagger$ Equal first authorship.},
 	Song Wu\textsuperscript{\rm 3}, 
 	Shuguang Cui\textsuperscript{\rm 1}
\affiliations
 	\textsuperscript{\rm 1} Shenzhen Research Institute of Big Data, The Chinese University of Hong Kong (Shenzhen), \\
 	\textsuperscript{\rm 2} Shanghai University,
 	\textsuperscript{\rm 3} Shenzhen Luohu Hospital

\emails
\textit{\{weibingzhao@link., xuyan1@link., lizhen@\}cuhk.edu.cn,
}}
\fi

\begin{document}

\maketitle

\begin{abstract}
   Point Cloud Sampling and Recovery (PCSR) is critical for massive real-time point cloud collection and processing since raw data usually requires large storage and computation.
   In this paper, we address a fundamental problem in PCSR: How to downsample the dense point cloud with arbitrary scales while preserving the local topology of discarding points in a case-agnostic manner (i.e. without additional storage for point relationship)?
   We propose a novel Locally Invertible Embedding for point cloud adaptive sampling and recovery (\textbf{PointLIE})\footnote{Our code is released through \url{https://github.com/zwb0/PointLIE}}.
   Instead of learning to predict the underlying geometry details in a seemingly plausible manner,
   PointLIE unifies point cloud sampling and upsampling to one single framework through bi-directional learning. 
   Specifically, PointLIE recursively samples and adjusts neighboring points on each scale. Then it encodes the neighboring offsets of sampled points to a latent space and thus decouples the sampled points and the corresponding local geometric relationship.
   Once the latent space is determined and that the deep model is optimized, 
   the recovery process could be conducted by passing the recover-pleasing sampled points and a randomly-drawn embedding to the same network through an invertible operation.
   Such a scheme could guarantee the fidelity of dense point recovery from sampled points. 
   Extensive experiments demonstrate that the proposed PointLIE outperforms state-of-the-arts both quantitatively and qualitatively.
\end{abstract}

\section{Introduction}

Recently, as a fundamental representation of 3D data, point cloud collected by various depth scanners or LiDAR sensors has been applied to diverse domains, such as autonomous driving~\cite{yan2020sparse}, cultural heritage reconstruction~\cite{xu2014tridimensional} and 3D immersive telepresence~\cite{orts2016holoportation}.
However, with the increasing capabilities of 3D data acquisition, gigabytes of raw point data can be generated per second (\textit{e.g.,} Velodyne HDL-64E can collect up to 2.2 million points per second). Therefore, it is untractable to process large-scale point clouds directly due to the huge demands for power consumption, computational cost and communication load. Point Cloud Sampling and Recovery (PCSR) task aims to sample meaningful points from dense point cloud to compress the scale of the original point cloud while preserving the local topology of discarding points for future reconstruction, which is critical for massive real-time point cloud collection and processing.

\begin{figure}[t]
		\centering
		\includegraphics[width=0.95\columnwidth]{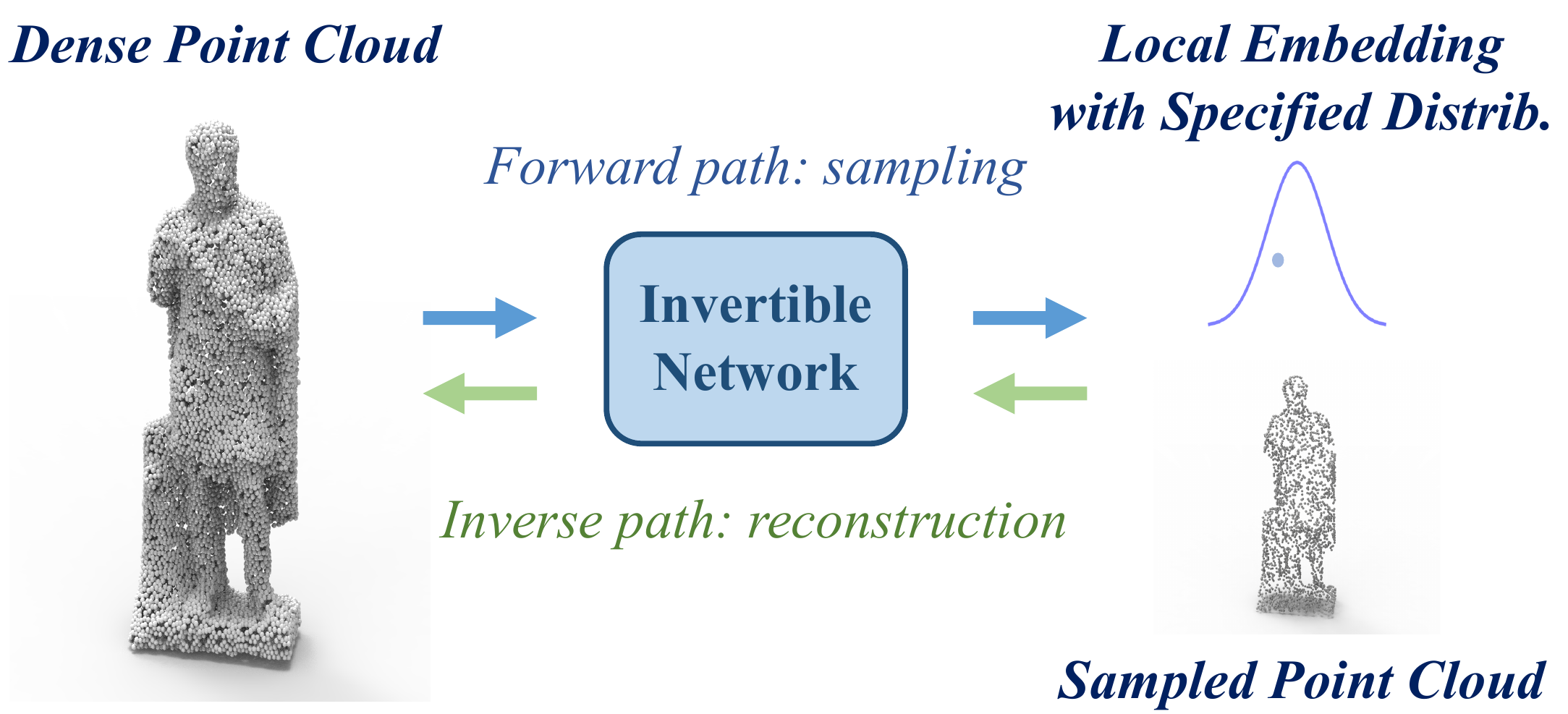} 
		\setlength{\belowcaptionskip}{-0.5cm}
		\caption{
		\textbf{Illustration of our PointLIE}. It incorporates the sampling and reconstruction processes into the same network, where the forward path transforms a dense point cloud to a sparse one and a case-agnostic latent variable.
		In the reverse path for reconstruction, a randomly drawn variable and the adaptively sampled point cloud are reconstructed to a dense one. 
		}
		\label{fig1}
	\end{figure}

In  point  cloud  research, significant  progress  has  been made in the single-track task for better compressing or upsampling the input point clouds, \textit{i.e.,} compressing point cloud with more pleasure surface approximations~\cite{schnabel2006octree,schwarz2018emerging} or upsampling  sparse  point cloud to dense point cloud~\cite{yu2018pu,yifan2019patch}.
However, bi-directional  PCSR remains challenging for several reasons: 
(1) Traditional methods compress point cloud through gradual sparseness. The global geometric configuration is lost in the intermediate process, rendering the intermediate results useless and invisible to downstream applications; 
(2) The relationship between data points requires huge storage.
(3) The simple up-sampling methods usually yield unsatisfactory recovering performance due to the loss of local structure information.
To deal with the above issues, we propose a novel scheme named  Locally Invertible Embedding (\textbf{PointLIE}) shown in Fig.~\ref{fig1} by sampling and upsampling point cloud for efficient storage and intermediate visibility, as well as complete and effective recovery.
In the down-scaling phase, PointLIE could generate viewable and recover-pleasing sub-point clouds with arbitrary scales while preserving the local offsets of discarding points recursively.
When point cloud recovery is needed, these sampled sub-points could be used to reconstruct the original dense point clouds more finely.
Inspired by the invertible neural network (INN) widely employed in both generative models~\cite{dinh2014nice,kingma2018glow,behrmann2019invertible,chen2019residual} and classification tasks~\cite{gomez2017reversible}, 
we design an INN-based learning framework to reduce the storage need for inter-points relationships.
To be more specific, it explicitly embeds the above local topology of missing points into a latent variable constrained to follow a specified distribution.
Therefore, it could generate a faithful dense point cloud simply by directly sampling a randomly-drawn latent variable and traversing the inverse pass of the network together with sampled sub-points.

In practice, PointLIE contains a Decomposition Module and several Point Invertible Blocks.
The former is adopted to transform the input point features into sampled point features and the offset residues for discarded points. 
After this, the number of point features is reduced to half, while the dimension of channels is expanded with higher-order offsets (\textit{i.e.,} subtracting point features from each other). 
By applying an elaborately designed cross-connection architecture, Point Invertible Blocks could further characterize the mutual interaction between the sampled features and their residue offsets in each sampling scale.
Following the aforementioned recursive process, the network could encode all offset information contained in every down-sampled scale into a locally invertible embedding.
In the training phase, we force such invertible embedding to conform to a pre-specified distribution (\textit{e.g.,} isotropic Gaussian) by using a distribution fitting loss.
Due to the complete reversible nature of PointLIE, the recovery process can be conducted by passing through PointLIE inversely, which is illustrated by the green arrow in Fig.~\ref{fig1}.

The main contributions of this paper are three folds. 
1) To the best of our knowledge, this is the first work that adopts INN in the PCSR task. 
A novel \textbf{PointLIE} scheme is proposed to model the sampling and upsampling stream into the same network through bi-directional learning.
2) We introduce a decomposition module and point invertible blocks to decouple the sampled point representations and the corresponding local neighbors' offsets in each down-sampled scale.
Meanwhile, a recursive invertible embedding is proposed to transform the offset of local neighbors into a latent variable that satisfies the specific distribution.
3) Extensive experiments demonstrate that PointLIE outperforms the state-of-the-art point cloud sampling and upsampling methods both quantitatively and qualitatively.

\section{Related Work}
\subsection{Sampling methods for Point Clouds}
Traditional sampling methods, such as Farthest point sampling (FPS), 
have wide applications in various point cloud frameworks~\cite{pointnet2,PointConv,Monte}, since they can sample relatively uniformly distributed points.
However, they do not take into account the subsequent processing of the sampled points and may result in sub-optimal performance.
Recently, there are some alternative sampling methods proposed to better capture the information of point clouds.
\cite{nezhadarya2020adaptive} introduced a critical points layer, which retains the critical points with the most active features to the next network layer.
\cite{Gumbel} proposed the Gumbel subset sampling using attention mechanisms to improve the classification and segmentation performance.
\cite{yan2020pointasnl} adaptively shifted the sampled points to objects' surface and thus increased the robustness of the network in noisy point clouds.
Other methods jointly consider sampling with downstream tasks. For example, \cite{dovrat2019learning,lang2020samplenet} introduced a task-specific sampling, which can improve the results through training with task-specific loss.
%
However, these methods improve the reconstruction mainly by joining the loss of specific tasks, while the geometric information lost in discarded points during sampling is not considered.

\begin{figure*}[t]
	\centering
	\includegraphics[width=0.9\textwidth]{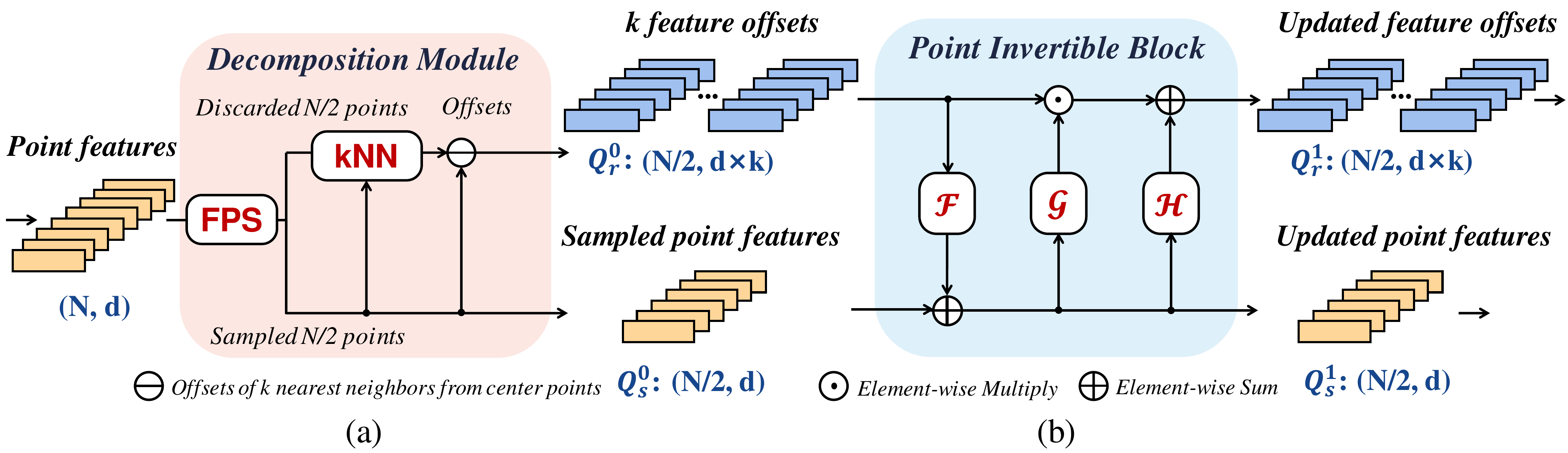} 
	\caption{\textbf{The internal structure of the Decomposition Module and Point Invertible Block.}
		(a) illustrates the Decomposition Module, which decouples the original point features into sampled point features and offsets to neighbor points features.
		(b) shows the Point Invertible (PI) Block, which updates the offsets and features of sampled points into their new counterparts.}
	\label{fig2}
\end{figure*}

\subsection{Upsampling methods for Point Clouds}

Point cloud upsampling aims to improve the point distribution density and uniformity.
%
%
\cite{yu2018pu} first proposed the neural network PU-Net, which learns point-wise features by PointNet++~\cite{pointnet2}, expanding the point set in feature space, and reconstructs an upsampled point set from those features.
%
%
3PU~\cite{yifan2019patch} is a multi-step progressive network, which learns different levels of detail in multiple steps.
%
However, due to its progressive nature, it requires a large amount of computation and more data to supervise the intermediate output of the network.
Recently, a Point Cloud Generative Adversarial Network (PU-GAN)~\cite{li2019pu} is designed to learn the distribution of the upsampled point set through adversarial learning.
Upsampling is an ill-posed problem since a downsampled point set corresponds to multiple plausible dense point clouds. Existing deep-learning based methods directly model this ambiguous task by learning the mapping from a sparse point set to a dense one under the supervision of the ground truth dense point set.
However, these methods fail to yield faithful complete reconstruction results, since the valuable information lost in the sampling process is ignored and irreversible.

\subsection{Invertible Neural Network}

Obtaining the measurable quantities (sampled points) from the given hidden parameters (sampling methods) is referred to as the forward process (\textit{i.e.,} sampling).
Correspondingly, the inverse process requires to infer the hidden states of a system from measurements (\textit{i.e.,} reconstruction). 
The inverse process is often intractable and ill-posed because valuable information is lost in the forward process~\cite{ardizzone2018analyzing}. 
To fully assess the diversity of possible inverse solutions for a given measurement, invertible neural networks (INNs) are employed to estimate the complete posterior of the parameters conditioned by observation, which is widely employed in both generative models~\cite{dinh2014nice,dinh2016density,kingma2018glow,behrmann2019invertible,chen2019residual} and classification tasks~\cite{gomez2017reversible,jacobsen2018revnet}.
Unlike traditional deep neural networks, which attempt to directly model the ambiguous problem of inferring the non-unique feasible result, INNs focus on learning the determinate forward process, using latent variables to capture the lost information. 
Due to the invertibility, the inverse process can be obtained for free by running through the network backwards.

\section{Methods}
\subsection{Task Overview}
Given a dense point set $\hat{\mathcal{Q}}=\{\hat{q}_i\}_{i=1}^{N}$, the goal of point cloud sampling and recovery (PCSR) with scale factor $r$ is to adaptively sample it into a sparse sub-point cloud $\mathcal{P}=\{p_j\}_{j=1}^{N/r}$ without any extra preservation, and then restore the dense point cloud $\mathcal{Q}=\{q_i\}_{i=1}^{N}$ from the sparse sub-point cloud.
To achieve the above goal, PointLIE is proposed as shown in Fig.~\ref{fig1}. 
The forward path decomposes the dense point cloud input into sampled points \P and a local invertible embedding $z$ containing the lost geometric information during sampling. 
Due to the reversible nature of PointLIE, the inverse path can reconstruct a faithful dense point cloud for free by running through the PointLIE backwards.
The whole process is formulated as,
\begin{align}
&f_{\theta}(\hat{\mathcal{Q}})=(\mathcal{P}, z), \text{ s.t. } z\sim p(z), \\
&f_{\theta}^{-1}(\mathcal{P},z^\star)=\mathcal{Q},~~ z^\star\sim p(z).
\label{0}
\end{align}
where $f_{\theta}(\cdot)$ denotes the forward path of our model, and $z$ is the local invertible embedding generated in the forward process, which is made to follow a specific distribution $p(z)$.
Note that here $z\sim p(z)$ is case-agnostic instead of case-specific ($z\sim p(z|\mathcal{P})$).
Therefore, there is no need to store $z$ after sampling, and we can just randomly draw an embedding $z^\star$ from the distribution $p(z)$ in the inverse path. \P and $z^\star$ are used to reconstruct a faithful \Q through the inverse process $f_{\theta}^{-1}(\cdot)$.



\begin{figure*}[t]
	\centering
	\includegraphics[width=\textwidth]{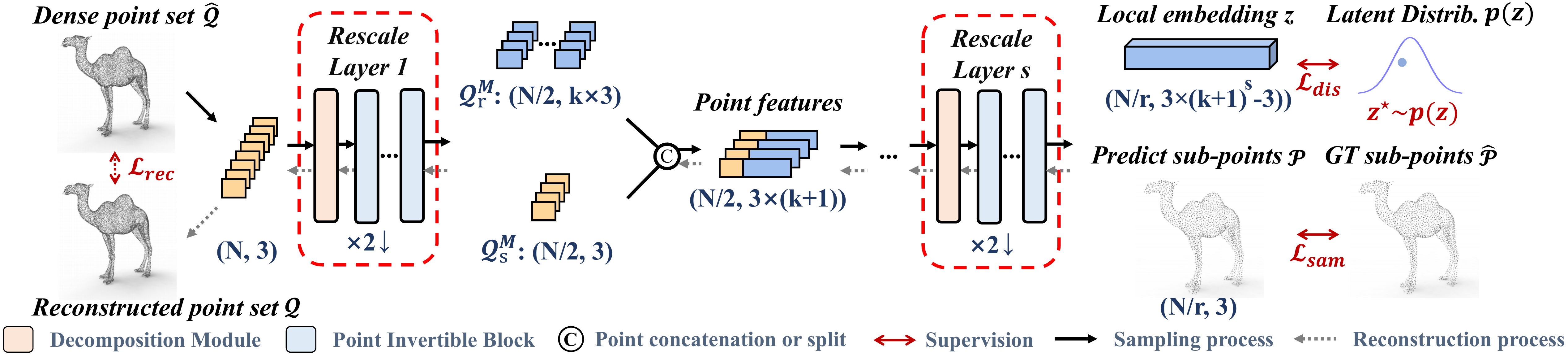} 
	\caption{\textbf{Illustration of the overall pipeline of PointLIE.} The black solid and grey dashed arrows indicate the flow of sampling and reconstruction respectively. 
	During the sampling path, the lost information represented by offsets to neighbor features are encoded into a latent variable $z$ following the distribution $p(z)$ by passing through $s$ Rescale Layers, where each Rescale Layer is composed of a Decomposition Module and several Point Invertible Blocks and samples the point set by half. 
	The reconstruction can be achieved through reverse operations.}
	\label{fig3}
\end{figure*}

\subsection{Invertible Architecture}
\label{arch}
To achieve the invertible operations, we firstly construct a \textit{Rescale Layer} by stacking a \textit{Decomposition Module} and $M$ \textit{Point Invertible Blocks} (PI Blocks) detailed in Fig.~\ref{fig2}.
When dealing with the PCSR with the scale factor $r$, we stack $s$ \textit{Rescale Layers} (s=$\lfloor\log_2r\rfloor$) to obtain the entire framework of PointLIE as illustrated in Fig.~\ref{fig3}. \\[.2cm]
\noindent\textbf{Decomposition Module.}
As shown in Fig.~\ref{fig2} (a), during the sampling process, the decomposition module is designed to separate the geometric information lost in discarded points from the sampled points.
Specifically, for input point features with the shape of $(N,d)$, we first conduct farthest point sampling (FPS) to select $N/2$ points while the remaining $N/2$ points are considered as discarded points.
To make the network preserve the information in discarded points, for each sampled point $q_i\in\mathcal{Q}_s$, we find its $k$ nearest neighbors in the discarded points and denote the spatial offsets from $q_i$ to its neighbors as $\mathcal{Q}_r$.
Here we use offsets rather than spatial coordinates of its neighbors since deep neural networks are more capable of learning the residues, and it is also easier to make the residues to follow the isotropic Gaussian distribution.
The decomposition module outputs two branches of features, \textit{i.e.,} $k$ feature offsets $\mathcal{Q}_r^0$ with shape $(N/2,d\times k)$ and sampled point features $\mathcal{Q}_s^0$ with shape $(N/2,d)$.\\[.2cm]
\noindent\textbf{Point Invertible Block.}
To further characterize the representation of the two branches during the forward path, 
we design a point invertible block to update features, inspired by the coupling layer in generative models~\cite{dinh2014nice,dinh2016density}.
As shown in Fig.~\ref{fig2} (b), each PI block takes two branches as input (\textit{i.e.,} the $1/2$ sampled point features $\mathcal{Q}_s^l$ and their kNN offsets $\mathcal{Q}_r^l$) and generates updated features $\mathcal{Q}_s^{l+1}$ and offsets $\mathcal{Q}_r^{l+1}$ by Eq.~\eqref{1}~\eqref{2}, 
%

\begin{align}
&\mathcal{Q}_s^{l+1} = \mathcal{Q}_s^l \odot \exp (\mathcal{Q}_r^l)+\mathcal{F}(\mathcal{Q}_r^l),\label{1} \\
&\mathcal{Q}_r^{l+1} = \mathcal{Q}_r^l\odot\exp(\mathcal{G}(\mathcal{Q}_s^{l+1}))+\mathcal{H}(\mathcal{Q}_s^{l+1}),\label{2}
\end{align}
where $l$ denotes passing through the $l$-th PI block, and $\mathcal{F}$, $\mathcal{G}$, $\mathcal{H}$ are three independent nonlinear transformations.
We use several stacked \texttt{conv1d} with the nonlinear activation for $\mathcal{F}$, $\mathcal{G}$, and the dense feature extractor in~\cite{yifan2019patch} for $\mathcal{H}$.
%
%
%
Note that PI blocks only enhance the representation of sampled features and neighboring offsets gradually, while the shapes of inputs and outputs of each PI block remain unchanged.\\[.2cm]
\noindent\textbf{Recursive Offset Residue Embedding.}
Fig.~\ref{fig3} illustrates the overall bi-directional pipeline of PointLIE for the PCSR task. 
By stacking $s$ rescale layers, where each of them contains a decomposition module and $M$ PI blocks, we construct a hierarchical structure for PCSR with arbitrary scales.
For each rescale layer, taking point features with the shape $(N,d)$ as an input, it will generate a $(N/2, d)$ sampled features and $(N/2,k\times d)$ feature offsets to $k$ neighbors.
Following this, a channel-dimension concatenation is conducted to merge the sampled features and their neighboring offsets to generate new point features.
These 'higher-order' point features will continue to be used as the input for the next rescale layer.
%
Therefore, the final embedding can be expanded to a series of high-order offsets recursively. $\mathcal{Q}_s^M$ and $\mathcal{Q}_r^M$ generated by the last Rescale Layer are treated as the adaptively sampled sub-point cloud \P and the embedding $z$. 
%
%
%
To fully illustrate the process of recursive offset embedding, we further present a special case for PCSR (\textit{i.e.,} including forward and inverse data stream) with $r=4$ and $k=1$ in the supplementary.\\[.2cm]
\noindent\textbf{Inverse Reconstruction Process.}
To reconstruct the original dense point set, we use the adaptively sampled point set \P and a randomly-drawn embedding $z^\star\sim p(z)$ as two branches of input to the reverse path of PointLIE (\textit{i.e.,} rescale layer $s$, $s-1$, ..., $1$) as indicated by the grey arrows in Fig.~\ref{fig3}. 
In each rescale layer, they will also flow in the reverse direction (\textit{i.e.,} PI block $M$, $M-1$, ..., $1$, decomposition module). 

The inverse operations of PI Blocks and the decomposition module are shown in Fig.~2 in the supplementary. In the reverse path, the $(l+1)$-th PI block 
aims to recover the neighboring offsets $\mathcal{Q}_r^l$ and the sampled features $\mathcal{Q}_s^l$ in the $l$-th block.
Considering the inputs $\mathcal{Q}_r^{l+1}, \mathcal{Q}_s^{l+1}$ with shapes $(N,k\times d)$ and $(N,d)$,
the reverse process of Eq.~\eqref{1}\eqref{2} can be expressed as,

\begin{align}
&\mathcal{Q}_r^l=(\mathcal{Q}_r^{l+1}-\mathcal{H}(\mathcal{Q}_s^{l+1}))\odot\exp(-\mathcal{G}(\mathcal{Q}_s^{l+1})),\label{3}\\
&\mathcal{Q}_s^l = (\mathcal{Q}_s^{l+1}-\mathcal{F}(\mathcal{Q}_r^l))\odot\exp(-\mathcal{Q}_r^l).
\label{4}
\end{align}
After reversely passing through $M$ PI blocks, the output $\mathcal{Q}_r^0, \mathcal{Q}_s^0$ will flow into the decomposition module reversely.
%
In detail, $\mathcal{Q}_r^0$ will be evenly split into $k$ offset matrices $\{Q_r^{(i)}\}_{i=1}^k$ along the channel dimension, where ${Q_r^{(i)}}$ with shape $(N,d)$ represents the offsets to the $i$-th nearest neighbour for each point in $\mathcal{Q}_s^0$ in the discarded points.
%
Then element-wise addition will be conducted between each ${Q_r^{(i)}}$ and $\mathcal{Q}_s^0$ respectively, obtaining features of recovered discarded points $\mathcal{Q}_d$ with shape $(kN,d)$.
$\mathcal{Q}_d$ will be concatenated with $\mathcal{Q}_s^0$ in a point-wise manner to form a candidate recovered point set $Q_c$ with shape $((k+1)\times N, d)$, whose first three dimensions in $d$ record the spatial coordinates. To guarantee the uniformity of the reconstructed points, we use FPS to select $2N$ point features from $Q_c$ based on their coordinates. 
%

Analogously, these $\times 2$ reconstructed point features will be evenly split into $(k+1)$ parts, where the first part and the remaining $k$ parts are taken as the sampled point features $\mathcal{Q}_s^M$ and the neighboring offsets $\mathcal{Q}_r^M$ respectively. 
%
Then, they will be fed into the next reversed rescale layer to conduct another $\times 2$ reconstruction.
Supported by Theorem 1, a faithful dense point cloud can be reconstructed progressively. The proof is provided in the supplementary.
%
%
%

%
\noindent\textbf{Theorem 1. } 
\textit{Suppose the generated invertible local embedding $z$ is subject to a latent distribution $p(z)$. 
	In the recovery process, by randomly sampling $z^\star$ from $p(z)$ and passing through the reverse path, the reconstructed dense point cloud $\mathcal{Q}$ will necessarily conform to the distribution of the real point cloud $p(\hat{\mathcal{Q}})$.}

\begin{table*}
	
	\caption{Performance comparison of PointLIE with state-of-the-arts for point cloud reconstruction. 
		\textbf{Bold} denotes the best performance.}
	\begin{center}
		
		\begin{tabular}{l|cc|ccc|ccc|ccc}
			\toprule[2pt]
			\multirow{2}*{Method}
			& Sampling & Network   &\multicolumn{3}{|c|}{Scale factor 4 ($10^{-3}$)}
			&\multicolumn{3}{|c}{Scale factor 8 ($10^{-3}$)} &\multicolumn{3}{|c}{Scale factor 16 ($10^{-3}$)}
			\\ ~ & mode & size (mean) & CD & HD & P2F & CD & HD & P2F & CD & HD & P2F\\
			\cline{2-10}
			\hline
			\hline
			PU-Net & FPS & 16.5 MB & 0.49 & 4.78 & 8.81 & 0.92 & 10.21 & 14.92 & 1.05 & 12.22 & 17.38 \\ 
			3PU & FPS & 92.5 MB & 0.41 & 4.86 & 2.72 & 0.54 & 8.91 & {3.68} &1.02 & 14.89 & \textbf{5.94}\\
			PU-GAN & FPS & 16.7 MB & 0.24 & 3.16 & {1.97} & 0.75 & 9.02 & 4.57 & 0.84 & 14.29 & 8.15 \\
			\hline
			PU-Net & SampleNet & 21.2 MB & 0.49 & 4.95 & 9.02 & 0.89 & 10.19 & 14.50 &1.04 &12.55 &  18.88\\ 
			PU-GAN & SampleNet & 21.4 MB & 0.23 & 2.89 & \textbf{1.93} &0.71 &8.03 & 4.57 & 0.80 &14.96 &8.06 \\
			
			PointLIE & - & 24.6 MB & \textbf{0.21} & \textbf{1.71} & {2.20} & \textbf{0.35} & \textbf{4.68} & \textbf{3.37} &\textbf{0.61} &\textbf{9.20} & {6.80}\\
			\toprule[2pt]
			
		\end{tabular}
	\end{center}
	\label{tab:tab1}
\end{table*}
\subsection{Training Objectives}
To improve the reconstruction result from an adaptively sampled point set, 
our PointLIE models the bi-directional transformation between the dense point cloud $\hat{\mathcal{Q}}$ and the sampled point cloud \P with a latent distribution $p(z)$.
Therefore, the total loss contains the following parts.\\[.2cm]
\noindent\textbf{Sparse Point Sampling Loss.}
Since the generated sampled point cloud \P is not the subset of $\hat{\mathcal{Q}}$,
we adopt the Earth Mover's distance loss (EMD)~\cite{fan2017point} $\mathcal{L}_{\text{sam}}$ to restrict \P to approach the 
original point cloud.\\[.2cm]
%
%
\noindent\textbf{Dense Point Reconstruction Loss.}
To reconstruct finer results, besides using EMD loss to restrict the geometric details of prediction, the reconstructed point set \Q should also be uniformly distributed on the surface of objects, thus repulsion loss $\mathcal{L}_{\text{rep}}$~\cite{yu2018pu} and uniform loss $\mathcal{L}_{\text{uni}}$~\cite{li2019pu} are used to distribute the recovered points \Q uniformly.
%
%
So the total loss for reconstruction is formulated as,

\begin{align}
&\mathcal{L}_{\text{rec}}=\lambda_{\text{emd}}\mathcal{L}_{\text{emd}}(\mathcal{Q},\hat{\mathcal{Q}})+\lambda_{\text{rep}}\mathcal{L}_{\text{rep}}+\lambda_{\text{uni}}\mathcal{L}_{\text{uni}}.
\end{align}
%
\noindent\textbf{Distribution Fitting Loss.}
Distribution fitting loss is used to encourage the distribution of the generated local embedding $f_\theta^z(\hat{\mathcal{Q}})$ to approach the latent distribution $p(z)$,
which is the sufficient condition for the reconstructed point set \Q to follow the real distribution of the original dense point set $\hat{Q}$ as proved in the Theorem 1.
In practice, the cross-entropy loss (CE) is employed to measure the difference between the distributions of the generated embedding $f_\theta^z(\hat{\mathcal{Q}})$ and $p(z)$. %
Here $p(z)$ is set as an isotropic Gaussian distribution, 
\begin{equation}
\begin{aligned}
\mathcal{L}_{\text{dis}}&=\text{CE}[f_{\theta}^z[p(\hat{Q})],p(z)]=-\mathbb{E}_{f_{\theta}^z[p(\hat{Q})]}[\log p(z)]\\
&=-\mathbb{E}_{p(\hat{Q})}[\log p(z=f_{\theta}^z(\hat{Q}))].
\end{aligned}
\label{ce}
\end{equation}
\noindent\textbf{Compound Loss.}
Overall, we train our PointLIE in an end-to-end manner by minimizing the total loss $\mathcal{L}$, 
\begin{equation}
\mathcal{L}=\lambda_{\text{sam}}\mathcal{L}_{\text{sam}}+\lambda_{\text{rec}}\mathcal{L}_{\text{rec}}
+\lambda_{\text{dis}}\mathcal{L}_{\text{dis}}.
\end{equation}

\section{Experiments}

\subsection{Dataset and Metrics}

To fully evaluate the proposed PointLIE, we compared our method with state-of-the-art methods on PU-147~\cite{li2019pu} dataset. 
This dataset integrates multiple objects from previous works (\textit{i.e.,} PU-Net~\cite{yu2018pu} and 3PU~\cite{yifan2019patch}, etc.), and ranges from simple and smooth models (\textit{e.g.,} icosahedron) to complex and high-detailed objects (\textit{e.g.,} statute).
We followed the official split of 120/27 for our training and testing sets.

During the experiment, we first sampled the input point cloud with different scale factors (\textit{i.e.,} 4, 8 and 16), and sampling modes (\textit{i.e.,} FPS and previous learnable sampling methods).
Then, we compared the reconstruction results with the ground truth point cloud.
For this propose, we used the Poisson disk sampling (PDS) method to uniformly sample 8192 points from each original mesh as our ground truth.
It should be noted that in our experiment, $16\times$ upsampling to 8192 points only took 512 points as input, 
which is more challenging than $16\times$ upsampling taking 5000 points as input in previous works~\cite{yifan2019patch,qian2020pugeo}.

To quantitatively evaluate the performance of different methods, we considered three commonly-used evaluation metrics, \textit{i.e.,} Chamfer distance (CD), Hausdorff distance (HD) and point-to-surface distance (P2F). 
The lower the metric values are, the better the reconstruction results are. 

\subsection{Implementation Details}
Under the premise of balancing efficiency and effectiveness, we set PI block number $M=8$ in the $4\times$ scale task, and $M=4$ in the rest $8\times$ and $16\times$ tasks. 
Furthermore, we set $k$ as $3$ to ensure that the information in the discarded points can be sufficiently preserved.
The details of the architecture will be shown in supplementary materials.



\begin{figure*}[t]
	\centering
	\includegraphics[width=0.95\textwidth]{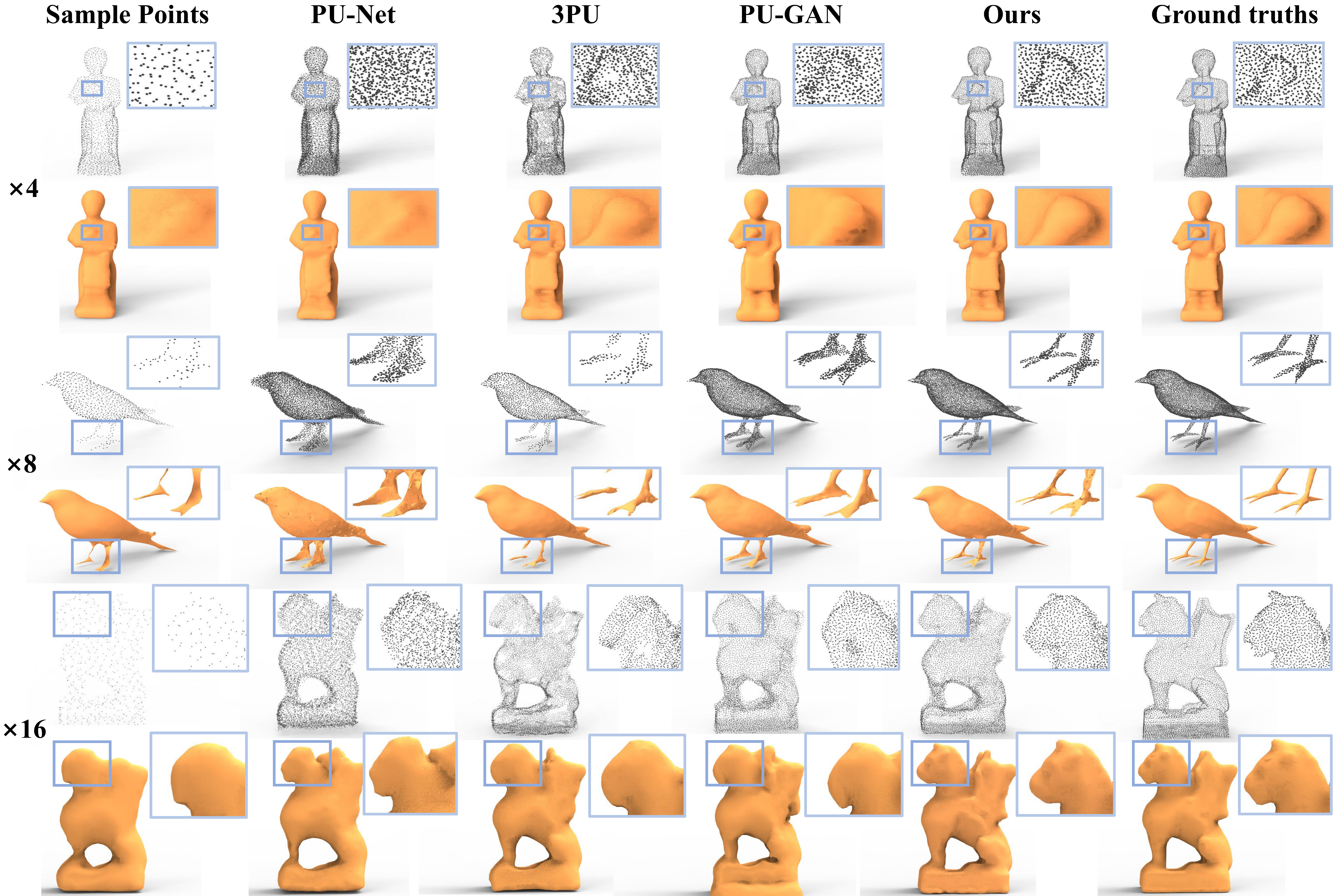} 
	\caption{Comparing the point set upsampling and surface reconstruction results produced by different methods.}
	\label{fig5}
\end{figure*}

\subsection{Quantitative Results}
\noindent\textbf{Reconstruct-guaranteed Point Cloud Sampling.}
In Tab.~\ref{tab:tab1} we compared the results of point cloud reconstruction with recent state-of-the-art methods: PU-Net~\cite{yu2018pu}, 3PU~\cite{yifan2019patch} and PU-GAN~\cite{li2019pu}.
To fairly compare with previous methods, we used different sampling modes to sample the input point clouds (\textit{i.e.,} FPS and adaptive sampling method~\cite{lang2020samplenet}). 
All methods were retrained with their public released codes on PU-147.

The upper part and the lower part of Tab.~\ref{tab:tab1} show the results by using FPS and the learnable sampling method respectively.
%
Among all, our PointLIE achieved the best results for most of the evaluation metrics,
especially for large scale PCSR tasks ($\times8$ and $\times16$).
Note that PointLIE achieves a slightly higher P2F value than previous methods in PCSR, partly because 
our sampling process adaptively adjusts the coordinates of the original points, deviating them slightly away from the surface.
However, we can achieve more visual-pleasing results in most cases especially for local geometric details.
Furthermore, we compared the results of the most appealing upsampling methods with state-of-the-art adaptive sampling methods (\textit{i.e.,} SampleNet).
We used the official codes of SampleNet~\cite{lang2020samplenet} and trained it with the upsampling downstream task by merging it with PU-Net and PU-GAN.
Here we did not use 3PU since it requires multiple SampleNet in training process.
The experiment results show that SampleNet cannot effectively improve the reconstruction performance.
Because PU-147 is more difficult than their experiment dataset ModelNet10 with only 10 fixed categories, and their network tends to over-memorize the properties of limited categories rather than acquire the local patterns.\\[.2cm]
\begin{table}
	\small
	\caption{Performance comparison of PointLIE with the state-of-the-arts for upsampling, where PU-GAN (-) indicates the results of PU-GAN without the discriminator.
		\textbf{Bold} denotes the best performance.
	}
	\begin{center}
		
		\begin{tabular}{l|cc|cc}
			\toprule[2pt]
			\multirow{2}*{Method}
			&\multicolumn{2}{|c|}{Scale 4 ($10^{-3}$)}
			&\multicolumn{2}{|c}{Scale 16 ($10^{-3}$)}
			\\ ~  & CD & HD & CD & HD\\
			\cline{2-5}
			\hline
			\hline
			PU-Net & 0.52 & 7.37 & 2.46 & 14.37\\ 
			3PU & 0.72 & 8.94 & 2.17 & 12.67\\
			PU-GAN (-) & 0.57 & 7.25 & 2.20 & 18.82 \\
			
			PU-GAN & \textbf{0.28} & \textbf{4.64} & 2.07 & 16.59\\
			\hline
			PointLIE & 0.32  & 4.93 & \textbf{1.98} & \textbf{12.08}\\
			\toprule[2pt]
			
		\end{tabular}
	\end{center}
	
	\label{tab:tab2}
\end{table}
\noindent\textbf{Point Cloud Upsampling.}
Our PointLIE can also be used as a general point cloud upsampling framework by feeding sparse input into the inverse stream of the trained model.
For a fair comparison with previous upsampling methods, we followed the experiment setting of~\cite{li2019pu}, feeding \textit{randomly} sampled 2048/512 points to predict 8192 dense output.
Tab.~\ref{tab:tab2} shows the quantitative comparison results with different appealing methods. 
Our PointLIE achieved comparable results in all evaluation metrics.
Particularly, our results by far exceed all previous methods without adversarial learning (\textit{e.g.,} PU-GAN (-)). PointLIE even outperforms complete PU-GAN in $\times 16$ task.
This result confirms that the performance improvement of PU-GAN mainly comes from the introduction of a discriminator rather than the model architecture itself,
while our architecture design can achieve superior upsampling results for both dense and sparse input. 
%

\subsection{Qualitative Results}
We also compared our qualitative results with PU-Net, 3PU and PU-GAN for point cloud reconstruction on different scales. Here PU-Net and PU-GAN took the points sampled by SampleNet as inputs.
Fig.~\ref{fig5} shows the visual results of point set recovery and surface reconstruction by~\cite{kazhdan2013screened}.
As shown in Fig.~\ref{fig5}, other methods tend to reconstruct more noisy and nonuniform point sets, resulting in more artifacts and ambiguities on the reconstructed surfaces.
Specifically, PointLIE generates more fine-grained details in the reconstructed results, especially for local geometric shapes (\textit{e.g.,} human hands, bird claws and dragon horns). 
More visualization results are shown in the supplementary.

\subsection{Ablation Study}
To further demonstrate the effectiveness of our proposed framework, we design an ablation study for different training modes and data feeding.
In Tab.~\ref{tab:tab3}, we first show the result produced without bi-directional learning (only training the inverse process) in the first two rows,
These results show that only using the inverse process during training cannot make the model learn the distribution of the reconstructed point cloud.
Then, we used the proposed training strategy mentioned in the experiment, which made a remarkable improvement dealing with randomly or uniformly sampled point clouds.
Finally, when we used the sub-pointset adaptively sampled by our network, further improvement is achieved.


\begin{table}
	\small
	\caption{Ablation study for PointLIE by using different training modes and input, where R and S refer to reconstruction and sampling process, respectively.
		\textbf{Bold} denotes the best performance.
	}
	\begin{center}
		
		\begin{tabular}{cc|c|cc}
			\toprule[2pt]
			Model & Sample mode & Bi-direction& CD & HD \\
			\hline
			\hline
			PointLIE-R & Random & \xmark & 4.93 & 16.59\\
			PointLIE-R & FPS & \xmark  & 2.32 & 7.58\\
			PointLIE-R  & Random & \cmark  & 0.32 & 4.93\\
			PointLIE-R  & FPS & \cmark  & 0.27 & {2.73}\\
			PointLIE-R & PointLIE-S  & \cmark  & \textbf{0.21} & \textbf{1.71} \\
			\toprule[2pt]
			
		\end{tabular}
	\end{center}
	\label{tab:tab3}
\end{table}

\subsubsection{Difference between point cloud upsampling.}
Although our PointLIE can yield satisfactory results on point cloud upsampling, our main focus lies in how to restore the point cloud for the subsequent tasks 
through a more suitable sampling method.
The PCSR task aims to restore the point cloud from an adaptively sampled sub-point cloud.

\subsubsection{Difference between point cloud compression.}
Our task is a special field of point cloud compression.
Previous point cloud compression preserves hidden results by invisible and meaningless latent codes, which partly harms the downstream process and inspection.
Inversely, our internal preservation is a sub-point cloud, which is fully viewable and available for the downstream tasks.

\section{Conclusion}

For the first time, we adopt the INN in the PCSR task and propose a completely new framework PointLIE, which models the sampling and upsampling streams into the same network through bi-directional learning. 
Different from the traditional point cloud compression, this framework can preserve visible results in the sampling process without preserving extra point relations.
By using one decomposition module and several point invertible blocks to decouple the sampled points with their local neighbors, our PointLIE can finely restore the original point cloud with a recursive invertible embedding using the reversed operations.
Extensive experiments demonstrate that PointLIE outperforms the state-of-the-art sampling and upsampling methods both quantitatively and qualitatively.

\section*{Acknowledgments}
The work was supported in part by the Key Area R\&D Program of Guangdong Province with grant No.2018B030338001, the National Key R\&D Program of China with grant No.2018YFB1800800, NSFC-Youth 61902335, Guangdong Regional Joint Fund-Key Projects 2019B1515120039, Shenzhen Outstanding Talents Training Fund, Shenzhen Institute of Artificial Intelligence and Robotics for Society, Guangdong Research Project No.2017ZT07X152 and CCF-Tencent Open Fund.
\newpage
\small
\bibliographystyle{named}
\bibliography{ref}

\newpage
\begin{center}
	{\textit{\LARGE\bf Supplementary Material}}
\end{center}

\thispagestyle{empty}
\setcounter{section}{0}
\setcounter{figure}{0}
\setcounter{table}{0}
\renewcommand\thesection{\Alph{section}}

\section{Overview}
In this supplementary material, we will first present a special case to clarify the recursive embedding process in the forward path for sampling in Section~\ref{Special Case}. In Section~\ref{Reverse Operations}, we illustrate the inverse flow within the Point Invertible Blocks and the Decomposition Module. Then we elaborate on the network architecture of our PointLIE and the implementation details in Section~\ref{Implementation Details}. 
In Section~\ref{Theoretical Proof}, we provide the theoretical proof of the Theorem 1 presented in the main paper. 
Then, we present additional experiments in Section~\ref{Additional Experiment}. We show the ablation studies on the architecture design, including the number of Point Invertible Layers in each Rescale Layer, the number of neighbors to find in the Decomposition Module and the scale of sampled $z$ from the isotropic Gaussian distribution. 
Also, in Section~\ref{Visualization Results}, we will provide more visualization results on  Point Cloud Sampling and Recovery (PCSR) tasks in different scales (\textit{i.e.,} 4$\times$, 8$\times$, 16$\times$) obtained by 1) the combination of FPS with upsampling methods (\textit{i.e.,} PU-Net~\cite{yu2018pu}, 3PU~\cite{yifan2019patch}, PU-GAN~\cite{li2019pu}); 2) Adaptive sampling network (\textit{i.e.,} SampleNet~\cite{lang2020samplenet}) and upsampling methods; 3) our proposed PointLIE. 
Besides, we will also compare the visualization results of the sub point cloud sampled by the Farthest Point Sampling (FPS) with ours sampled by PointLIE.
Finally yet importantly, the PCSR results on real-scanned LiDAR point clouds (from KITTI dataset~\cite{geiger2013vision}) generated by our PointLIE will also be presented.
\section{Special Case}
\label{Special Case}
\begin{figure}[htbp]
	\centering
	\includegraphics[width=0.9\linewidth]{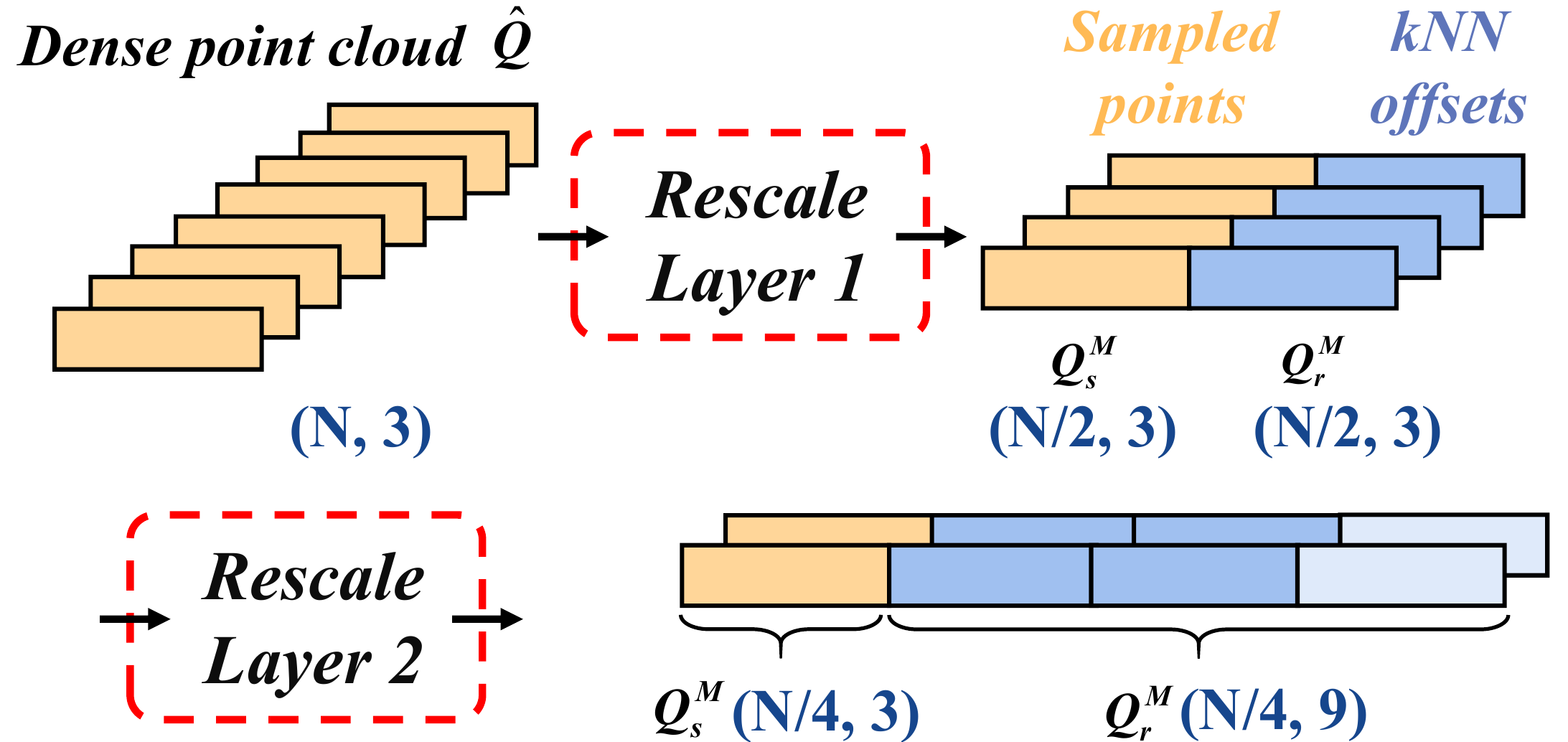} 
	\caption{Special case of recursive invertible embedding for $\times 4$ PCSR with $k=1$ (\textit{i.e.,} selecting one neighbour).
		Yellow indicates the coordinates, while blue indicates offsets (lighter blue indicates higher-order offsets).}
	\label{fig:case}
\end{figure}

To fully illustrate the recursive embedding process of PointLIE, we further illustrate a case of PCSR with scale factor 4 and $k=1$ in Fig.~\ref{fig:case}. After each Rescale Layer, the number of points are reduced to half, and the dimension of kNN offset features is expanded progressively.

\begin{figure*}[t]
	\centering
	\includegraphics[width=\textwidth]{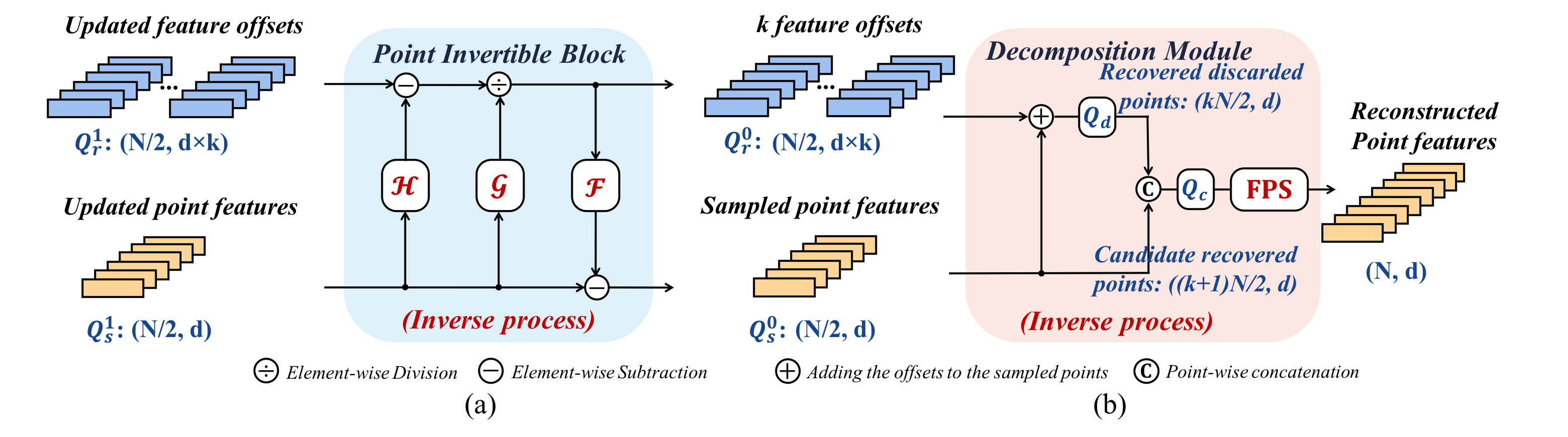} 
	\caption{\textbf{The reverse flow of the Point Invertible Block and the Decomposition Module.}
		(a) illustrates the inverse path of Point Invertible (PI) Block, which recovers the offset features $\mathcal{Q}_r^l$ and sampled point features $\mathcal{Q}_s^l$ from $\mathcal{Q}_r^{l+1}$ and $\mathcal{Q}_s^{l+1}$, where $l$ denotes the $l$-th PI Block.
		(b) shows the inverse path of the Decomposition Module, which adds the offsets to their corresponding sampled points to reconstruct $(k+1)N/2$ candidate points $\mathcal{Q}_c$, followed by FPS to select $N$ of them as the $\times 2$ points reconstructed by this Rescale Layer.
	}
	\label{fig:inverse}
\end{figure*}
\section{Reverse Operations}
\label{Reverse Operations}
Please refer to the Sec.~'Inverse Reconstruction Process' in the main text for better understanding.
To clarify the reverse reconstruction process, Fig.~\ref{fig:inverse} further shows the detailed inverse operations of the two main components in the Rescale Layer, \textit{i.e.,} the Point Invertible Block and the Decomposition Module. In the reverse path for reconstruction, the adaptively sampled point cloud \P and a latent variable $z^\star$ randomly drawn from $p(z)$ are taken as the inputs and flow into $s$ Rescale Layers reversely to reconstruct $\times 2^s$ points as shown in the Fig.~3 in the main text. 

Within each Rescale Layer, the offset features $\mathcal{Q}_r^{l+1}$ and sampled point features $\mathcal{Q}_s^{l+1}$ will first flow into several Point Invertible Blocks to recover $\mathcal{Q}_r^l$ and $\mathcal{Q}_s^l$, where $l$ denotes the $l$-th PI Block. After passing through $M$ PI Blocks, $\mathcal{Q}_r^0$ and $\mathcal{Q}_s^0$ will flow into the Decomposition Module to generate $\times 2$ reconstructed points features. 
In detail, $\mathcal{Q}_r^0$ will be evenly split into $k$ offset matrices $\{Q_r^{(i)}\}_{i=1}^k$ along the channel dimension, where ${Q_r^{(i)}}$ with shape $(N/2,d)$ represents the offsets to the $i$-th nearest neighbour for each point in $\mathcal{Q}_s^0$ in the discarded points.
Then element-wise addition will be conducted between each ${Q_r^{(i)}}$ and $\mathcal{Q}_s^0$ respectively, obtaining features of recovered discarded points $\mathcal{Q}_d$ with shape $(kN/2,d)$.
$\mathcal{Q}_d$ will be concatenated with $\mathcal{Q}_s^0$ in a point-wise manner to form a candidate recovered point set $Q_c$ with shape $((k+1)N/2, d)$, whose first three dimensions in $d$ record the spatial coordinates. To guarantee the uniformity of the reconstructed points, we use FPS to select $N$ point features from $Q_c$ based on their coordinates.

\section{Implementation Details}
\label{Implementation Details}
During the experiment, we set PI block number $M=8$ in the $4\times$ scale PCSR task, and $M=4$ in the rest $8\times$ and $16\times$ tasks. 
Furthermore, we set $k$ as $3$ to ensure that the information in the discarded points can be sufficiently preserved.
In each Invertible block, $\mathcal{F}$, $\mathcal{G}$, $\mathcal{H}$ are three independent nonlinear transformations.
We use several stacked \texttt{conv1d} with the nonlinear activation for $\mathcal{F}$, $\mathcal{G}$. 
To be more specific, it uses two conv layers with nonlinear activation function to update point-wise features.
Then, a global pooling in all points is conducted to aggregate the global features.
Finally, another two-layer conv layer fine-tunes the feature and generates the updated features.
For $\mathcal{H}$, we use the dense feature extractor in~\cite{yifan2019patch} while slightly reducing the amount of parameters.
Here, we just use two DenseConv layers. Each layer will concatenate the features of updated features and all previous features.

During the training, we strictly followed the same training protocols as other methods for a fair comparison, augmenting the network input by random rotation, scaling, and point perturbation with Gaussian noise. 
We trained the network for 30 epochs using the Adam algorithm with the batch size of 6.
The learning rate of the network is initialized as 0.001 and dropped 0.3 for every 50k iteration until $10^{-6}$.

In the evaluation phase, we followed the commonly used patch-based prediction (\textit{i.e.,} \cite{li2019pu,yu2018pu,yifan2019patch}) in the inverse phase and cropped small patches around uniformly selected seeds.
Then we merged the upsampled patches and conducted FPS sampling to obtain the final reconstructed objects.
All experiments are implemented with TensorFlow and a single NVIDIA Titan Xp GPU.

\section{Theoretical Proof}
\label{Theoretical Proof}
\noindent\textbf{Theorem 1. } 
\textit{Suppose the generated invertible local embedding $z$ is subject to a latent distribution $p(z)$. 
	In the reconstruction process, by randomly sampling a $z^\star$ from $p(z)$ and passing it through the reverse path, the reconstructed dense point cloud $\mathcal{Q}$ will necessarily conform to the distribution of the real point cloud $p(\hat{\mathcal{Q}})$.}

\noindent\textbf{Proof.}
Since the real dense point set $\hat{\mathcal{Q}}$ follows the distribution $p(\hat{\mathcal{Q}})$, the adaptively sampled sub-point cloud \P and the generated local embedding $z$ also form their specific distributions respectively. We simply denote the joint distribution $p(\mathcal{P},z)$ as $f_{\theta}[p(\hat{Q})]$, where $f_{\theta}^\mathcal{P}[p(\hat{Q})]$ and $f_{\theta}^z[p(\hat{Q})]$ represent the distribution of the sampled sub-point cloud and that of our generated local embedding respectively.
The reconstructed point cloud \Q follows the distribution
$f_{\theta}^{-1}[p(\mathcal{P}, z)]$.
Since the sub point cloud $\mathcal{P}$ and the latent variable $z$ are independent, 
%
the distribution of the reconstructed point cloud \Q can be expressed as,
\begin{equation}
\begin{aligned}
p(\mathcal{Q})&=f_{\theta}^{-1}[p(\mathcal{P}, z)]\xlongequal{\text{Ind.}}f_{\theta}^{-1}[p(\mathcal{P})p(z)]\\
&=f_{\theta}^{-1}[f_{\theta}^\mathcal{P}[p(\hat{Q})]p(z)]\label{pQ}
\end{aligned}
\end{equation}

Therefore, if $f_{\theta}^z[p(\hat{Q})]$ is forced to obey a pre-defined distribution $p(z)$ (\textit{e.g.,} an isotropic Gaussian distribution), \textit{i.e.,} $\forall \epsilon>0$, if $\mathcal{D} [f_{\theta}^z[p(\hat{Q})] , p(z) ] < \epsilon$, it has,
\begin{equation}
\begin{aligned}
&\mathcal{D} [f_{\theta}^z[p(\hat{Q})] , p(z) ] < \epsilon \label{condition}\\
\xLongrightarrow{\times f_{\theta}^\mathcal{P}[p(\hat{Q})]}
&\mathcal{D}[ f_{\theta}^\mathcal{P}[p(\hat{Q})]f_{\theta}^z[p(\hat{Q})], f_{\theta}^\mathcal{P}[p(\hat{Q})]p(z)]< \epsilon\\
\xLongrightarrow{\text{Ind.}}
&\mathcal{D}[ f_{\theta}^\mathcal{P}[p(\hat{Q})]p(z),f_{\theta}[p(\hat{Q})]]< \epsilon\\
\xLongrightarrow{\text{Backward}} &\mathcal{D}[ f_{\theta}^{-1}[f_{\theta}^\mathcal{P}[p(\hat{Q})]p(z)],p(\hat{Q})]< \epsilon\\
\xLongrightarrow{\text{Eq.~\eqref{pQ}}}  &\mathcal{D}[ p(\mathcal{Q}),p(\hat{\mathcal{Q}})]< \epsilon,
\end{aligned}
\end{equation}

where $\mathcal{D}$ is used to measure the difference between two distributions.

\section{Additional Experiment}
\label{Additional Experiment}

In this section, we discuss the settings of different components in our PointLIE.
Tab.~\ref{sup:tab:tab1} illustrates different model settings, where Model A and B show the results of different block number $M$. Model C and D show the influence of the number of neighbors $k$. Model E and F explore different scales of the latent distribution. 
In the last row of Tab.~\ref{sup:tab:tab1}, Model G with the default setting in the manuscript achieves the best result.

\begin{table}
	\small
	\caption{Ablation studies for PointLIE in $\times 4$ PCSR task by using different hyper-parameters, where $M$, $k$ and $GS$ refer to the number of invertible blocks, the number of neighbors in decomposition module and the scale of Gaussian distribution respectively.
		\textbf{Bold} denotes the best performance.
	}
	\begin{center}
		
		\begin{tabular}{c|ccc|cc}
			\toprule[2pt]
			Model & $M$ & $k$ & $GS$ & CD & HD \\
			\hline
			\hline
			A & 4 & 3 & 1.0 & 0.31 & 3.77\\
			B & 6 & 3  & 1.0 & 0.39 & 2.89\\ \hline
			C  & 8 & 1  & 1.0 & 0.25 & 2.23\\
			D & 8 & 5 & 1.0 & 0.31 & 2.68 \\\hline
			E  & 8 & 1  & 0.5 & 0.23 & 2.18\\
			F & 8 & 5 & 2.0 & 0.78 & 14.25 \\ \hline
			G  & 8 & 3  & 1.0 & \textbf{0.21} & \textbf{1.71}\\
			\toprule[2pt]
			
		\end{tabular}
	\end{center}
	\label{sup:tab:tab1}
\end{table}

\section{Visualization Results}
\label{Visualization Results}
\begin{figure*}[t]
	\centering
	\includegraphics[width=\textwidth]{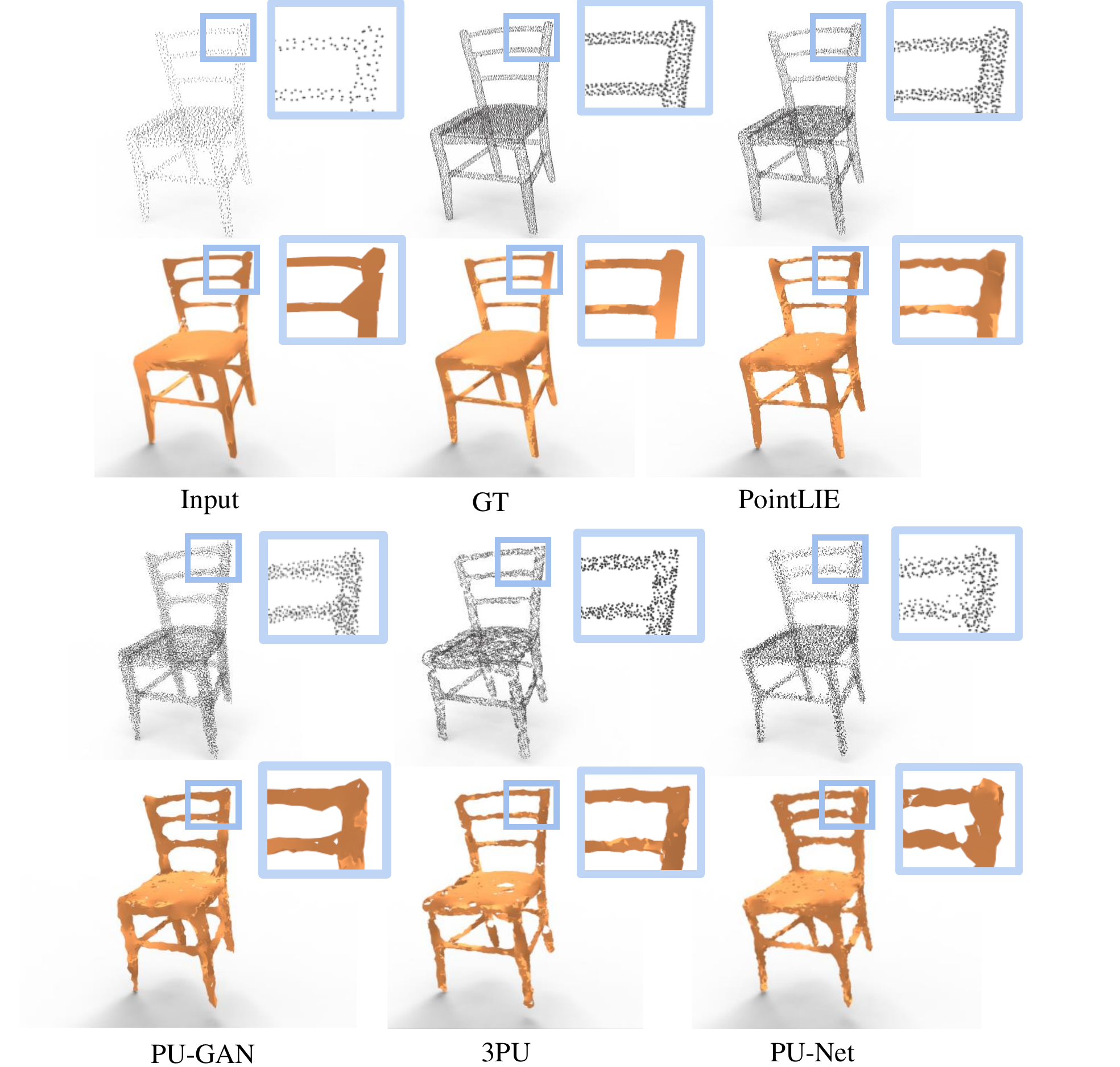} 
	\caption{Comparing the $\times 4$ PCSR and surface reconstruction results produced by different methods.}
	\label{sup:fig1}
\end{figure*}

In this section, we will provide more visualization results of our sampled points (see Fig.~\ref{sup:fig4}), PCSR for man-made (see Fig.~\ref{sup:fig1},\ref{fig2},\ref{sup:fig3}) and real-scanned point clouds (see Fig.~\ref{sup:fig5}).

\begin{figure*}[t]
	\centering
	\includegraphics[width=\textwidth]{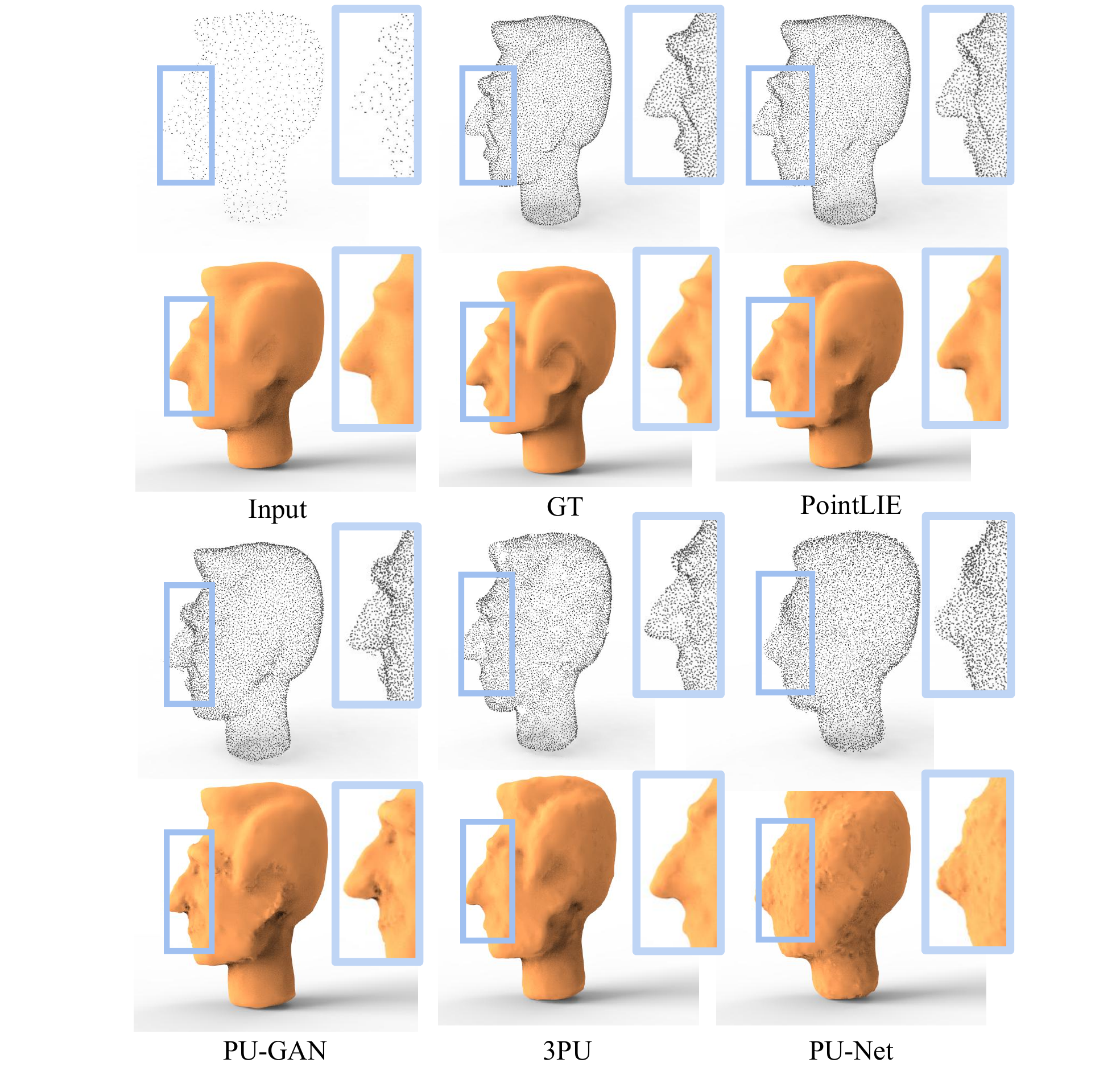} 
	\caption{Comparing the $\times 8$ PCSR and surface reconstruction results produced by different methods.}
	\label{sup:fig2}
\end{figure*}

\begin{figure*}[t]
	\centering
	\includegraphics[width=\textwidth]{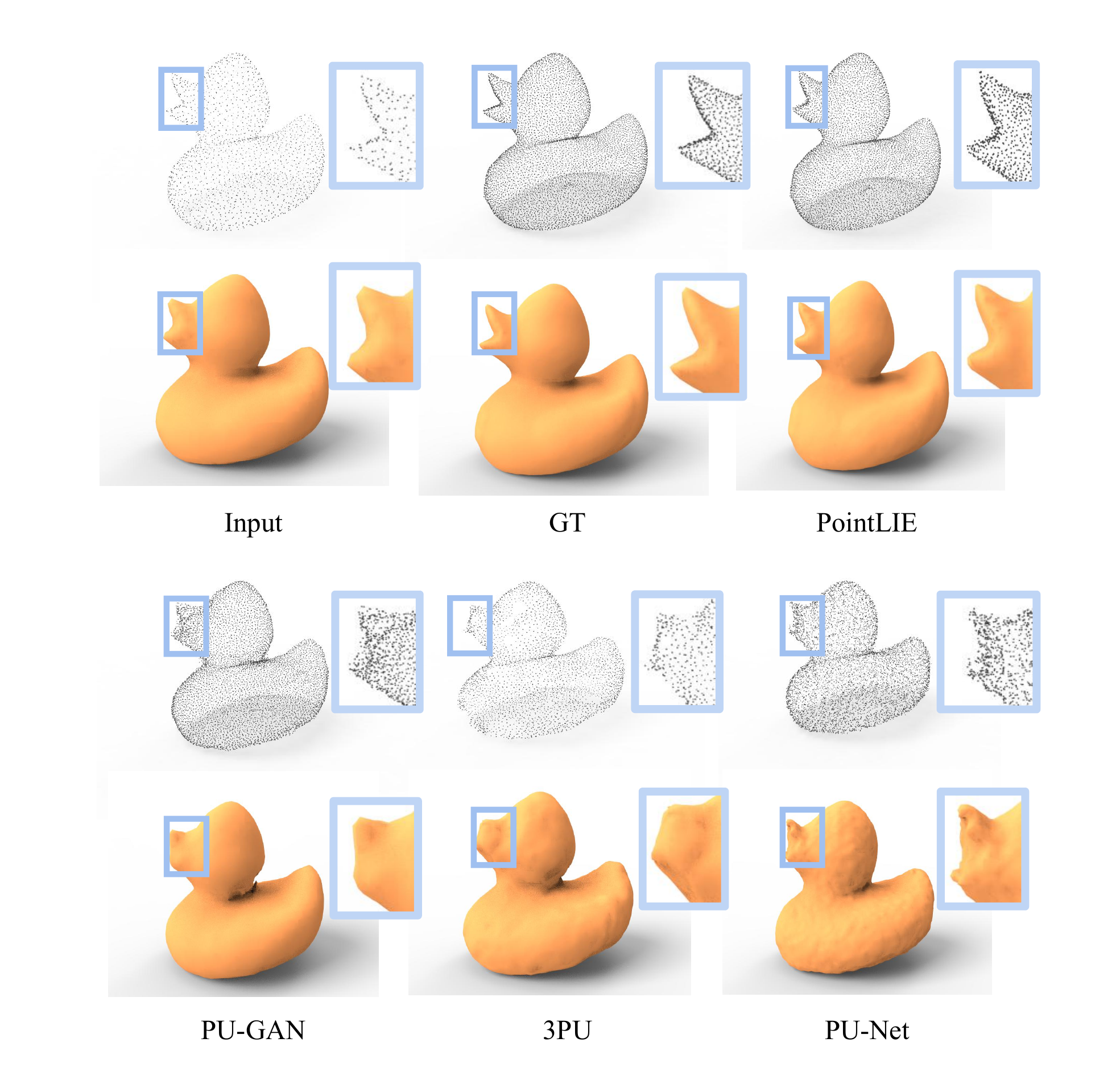} 
	\caption{Comparing the $\times 16$ PCSR and surface reconstruction results produced by different methods.}
	\label{sup:fig3}
\end{figure*}

\begin{figure*}[t]
	\centering
	\includegraphics[width=\textwidth]{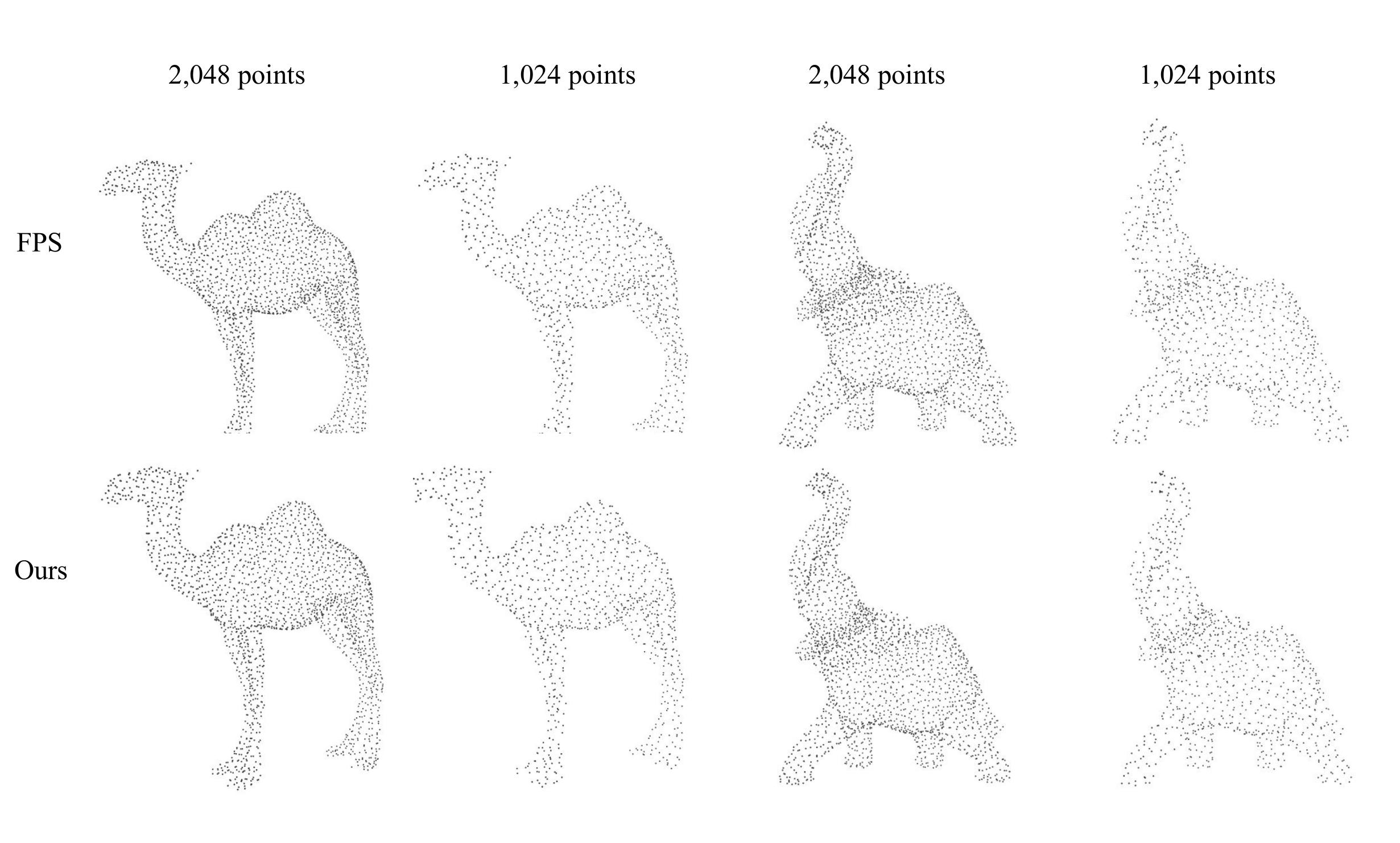} 
	\caption{Comparing our sampled point cloud with FPS. The point set adaptively sampled by PointLIE not only resembles the point set uniformly sampled by FPS but also tends to cover the detailed structures, which is beneficial for reconstruction and subsequent tasks.}
	\label{sup:fig4}
\end{figure*}

\begin{figure*}[t]
	\centering
	\includegraphics[width=\textwidth]{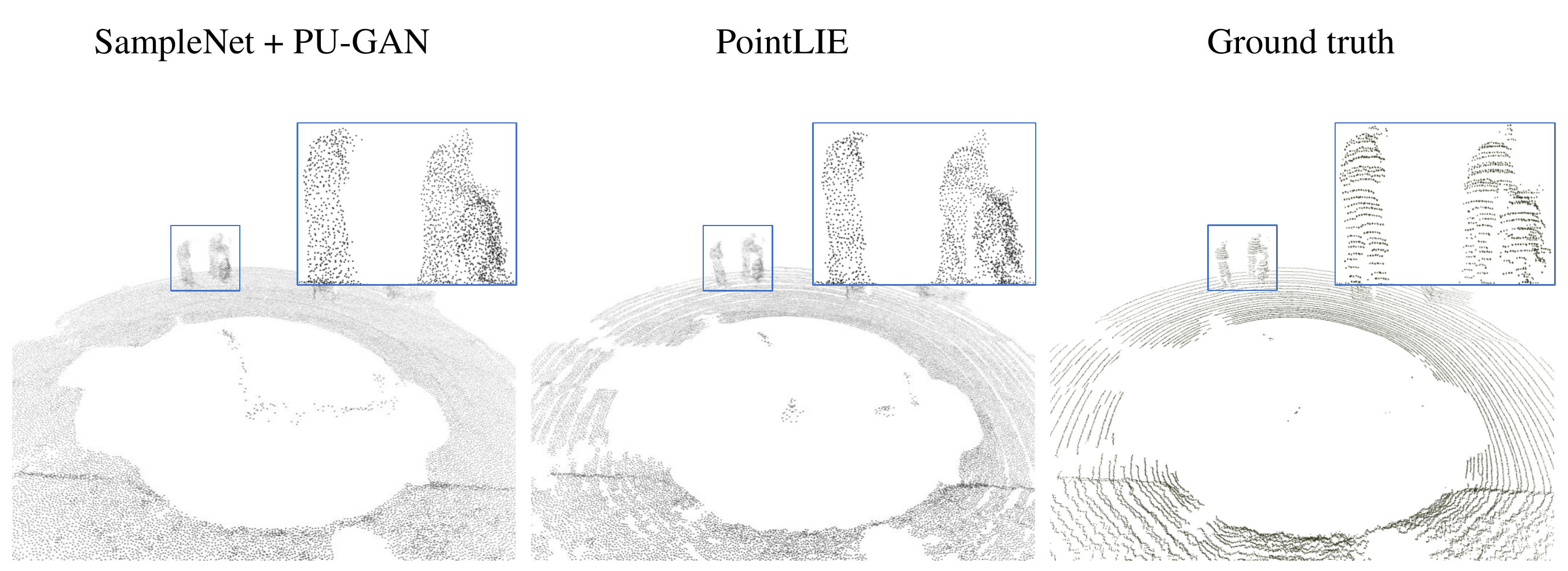} 
	\caption{Real-scanned results. Comparing SampleNet+PU-GAN with PointLIE for PCSR task on real-scanned large scale LiDAR point cloud, it can be observed that our PointLIE can recover more realistic spatial relationship compared with SampleNet+PU-GAN (\textit{e.g.,} the gap between the pedestrian and motorbike).}
	\label{sup:fig5}
\end{figure*}

Fig.~\ref{sup:fig1},\ref{sup:fig2},\ref{sup:fig3} further present more visual comparison results by applying our PointLIE and SampleNet combined with previous state-of-the-art methods, (\textit{i.e.,} PU-Net~\cite{yu2018pu}, 3PU~\cite{yifan2019patch}, PU-GAN~\cite{li2019pu}) on the reconstruction-guaranteed point cloud sampling (PCSR) task with scale factors $\times 4, \times 8$ and $\times 16$ respectively. The surface reconstruction results are also presented for better comparison. From these visualization results, it can be observed that our PointLIE can generate a more faithful point cloud, which is more consistent with the dense ground truth point cloud and preserve more detailed structures (\textit{i.e.,} the rails of chair, the nose of sculpture and the beak of duck).

Fig.~\ref{sup:fig4} shows the visual results of the sub-point set sampled by FPS and by PointLIE. The point set adaptively sampled by PointLIE not only resembles the point set uniformly sampled by FPS but also tends to cover the detailed structures, which is beneficial for reconstruction and subsequent tasks.

Besides, we also present some PCSR results on real-scanned LiDAR point clouds as shown in Fig.~\ref{sup:fig5}. The point set recovered by the PointLIE preserves more detailed information. It can be observed that our PointLIE can recover more realistic spatial relationship compared with SampleNet+PU-GAN (\textit{e.g.,} pedestrian and motorbike), which may partially result from the plausible GAN loss used in PU-GAN.  

\end{document}